\documentclass[letterpaper, 10 pt, conference]{ieeeconf}  % Comment this line out if you need a4paper

\IEEEoverridecommandlockouts                              % This command is only needed if 
\UseRawInputEncoding

\usepackage{graphics}
\usepackage{epsfig}
\usepackage{mathptmx}
\usepackage{times} 
\usepackage{amssymb}  
\usepackage{multirow}
\usepackage{longtable}
\usepackage{amsmath}
\usepackage{lipsum,lineno}
\usepackage{float}
\usepackage{url}
\usepackage{mathtools}
\usepackage{algorithm}
\usepackage{algpseudocode}
\usepackage{changepage}
\usepackage{hyperref}
\usepackage{caption}
\usepackage{subcaption}
\usepackage{amsmath}
\usepackage{soul} %sout
\usepackage{amsfonts}

\let\labelindent\relax
\usepackage{enumitem}

\usepackage{eucal} % for stylish mathcal
\usepackage{booktabs} 
\usepackage{adjustbox} % for tables
\usepackage[normalem]{ulem}
\useunder{\uline}{\ul}{}
\usepackage{multirow}
\usepackage{graphicx}
\usepackage[table,xcdraw]{xcolor}

\newcommand{\obj}[1]{o_{#1}}

\newcommand{\obje}[1]{\hat{o}_{#1}}
\newcommand{\objn}[1]{o_{n}}
\newcommand{\cam}[1]{c_{#1}}
\newcommand{\taskf}[1]{t_{#1}}
\newcommand{\world}[1]{w_{#1}}
\newcommand{\gripperf}[1]{g_{#1}}
\newcommand{\basef}[1]{b_{#1}}

\newcommand{\error}[0]{\epsilon}

\newcommand{\grasp}[0]{g}
\newcommand{\baseik}[0]{b}

\newcommand{\transform}[2]{\mathbf{T}_{#2}^{#1}}

\newcommand{\prob}[0]{\mathrm{P}}

\newcommand{\poseestimate}[1]{\hat{X}_{#1}}
\newcommand{\poseuncertainty}[1]{\prob(X_{#1})}

\newcommand{\Tob}[0]{\transform{\obj{}}{\baseik}}
\newcommand{\Tog}[0]{\transform{\obj{}}{\grasp}}
\newcommand{\Tot}[0]{\transform{\obj{}}{\taskf{}}}

\newcommand{\Twt}[0]{\transform{\world{}}{\taskf{}}}
\newcommand{\Tco}[0]{\transform{\cam{}}{\obj{}}}
\newcommand{\Twc}[0]{\transform{\world{}}{\cam{}}}

\newcommand{\Tcoe}[0]{\transform{\cam{}}{\obje{}}}

\newcommand{\Toeo}[0]{\transform{\obje{}}{\obj{}}}
\newcommand{\Tono}[0]{\transform{\objn{}}{\obj{}}}
\newcommand{\Toon}[0]{\transform{\obj{}}{\objn{}}}

\newcommand{\baseposeest}[0]{\transform{\obje{}}{\baseik}}
\newcommand{\graspposeest}[0]{\transform{\obje{}}{\grasp}}

\newcommand{\samplecont}[0]{x}

\newcommand{\errorspace}[0]{\mathcal{E}}
\newcommand{\eacc}[0]{\mathcal{E}_{\text{acc}}^{\obje{}}}
\newcommand{\eest}[0]{\mathcal{E}_{\text{est}}^{\obje{}}}

\newcommand{\errorn}[0]{\epsilon_{n}}
\newcommand{\ecomp}[2]{\epsilon_{#1_#2}}
\newcommand{\ein}[2]{N_{#1_#2}}
\newcommand{\tdf}[0]{\delta t}
\newcommand{\rdf}[0]{\delta r}
\newcommand{\task}[0]{\text{task}}

\DeclareMathOperator*{\argmin}{argmin}

\title{\LARGE \bf Robotic Task Success Evaluation Under Multi-modal Non-Parametric Object Pose Uncertainty}

\author{Lakshadeep Naik$^{1}$,  Thorbj\o rn Mosekj\ae r Iversen$^{1}$, Aljaz Kramberger$^{1}$, and Norbert Kr\"uger$^{1,2}$%<-this % stops a space
\thanks{$^{1}$SDU Robotics, M\ae rsk Mc-Kinney M\o ller Institute (MMMI),  Faculty of Engineering, University of Southern Denmark, Odense M, Denmark {\tt\small lana,thmi,alk,norbert@mmmi.sdu.dk}}%
\thanks{$^{2}$Danish Institute for Advanced Studies (DIAS), Odense M, Denmark}%
}
\begin{document}

\maketitle
\thispagestyle{empty}
\pagestyle{empty}

%%%%%%%%%%%%%%%%%%%%%%%%%%%%%%%%%%%%%%%%%%%%%%%%%%%%%%%%%%%%%%%%%%%%%%%%%%%%%%%%

\begin{abstract}
Accurate 6D object pose estimation is essential for various robotic tasks. Uncertain pose estimates can lead to task failures; however, a certain degree of error in the pose estimates is often acceptable. Hence, by quantifying errors in the object pose estimate and acceptable errors for task success, robots can make informed decisions. This is a challenging problem as both the object pose uncertainty and acceptable error for the robotic task are often multi-modal and cannot be parameterized with commonly used uni-modal distributions. In this paper, we introduce a framework for evaluating robotic task success under object pose uncertainty, representing both the \textit{estimated error space} of the object pose and the \textit{acceptable error space} for task success using multi-modal non-parametric probability distributions. The proposed framework pre-computes the \textit{acceptable error space} for task success using dynamic simulations and subsequently integrates the pre-computed \textit{acceptable error space} over the \textit{estimated error space} of the object pose to predict the likelihood of the task success. We evaluated the proposed framework on two mobile manipulation tasks. Our results show that by representing the \textit{estimated} and the \textit{acceptable} \textit{error space} using multi-modal non-parametric distributions, we achieve higher task success rates and fewer failures.
\end{abstract}

\section{INTRODUCTION}
\label{sec:intro}

\begin{figure}[]
    \centering
    \includegraphics[width=1.00\columnwidth]{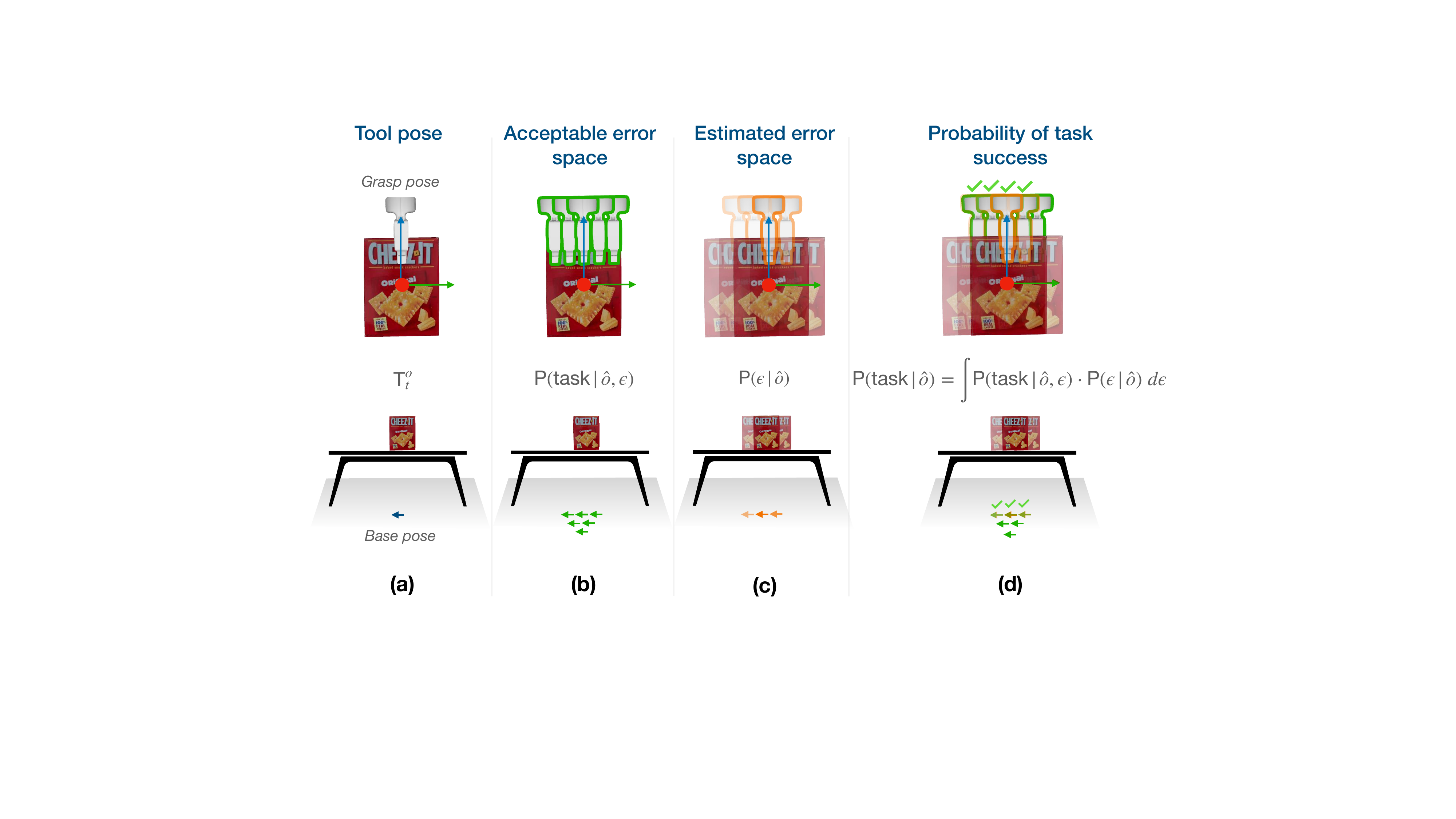}
    \caption{\textbf{a)} Selected tool pose (grasp and base pose) relative to the object frame $\obj{}$
    % , $\Tot$. 
    \textbf{b)} \textit{Acceptable error space} 
    % $\eacc$ 
    pre-computed by introducing errors $\error$ in object frame $\obj{}$ and determining the probability of task success 
    % given the error $\error$ in $\obj{}$
    % , $\prob(\task|\obj{},\error)$ 
    \textbf{c)} \textit{Estimated error space} 
    % $\eest$ 
    for the object frame estimate $\obje{}$ obtained using estimated visual object pose distribution \textbf{d)} The probability of task success is calculated by integrating the \textit{acceptable error space} over the \textit{estimated error space}.}
    \label{fig:idea}
\end{figure}

% 6D object pose for robotic tasks and examples
Many robotic tasks, such as grasping rigid objects, require a 6D pose estimate of the object. For example, to grasp the object using a mobile manipulator, first, suitable base poses and grasp poses are selected relative to the object (see Fig~\ref{fig:idea}a). The estimate of both poses is obtained using the estimated 6D pose of the object. Consequently, any error in the estimated object pose also propagates to the estimated base or grasp pose, potentially leading to task failure.

% why delaying the action better than failing
Failures can incur significant costs in robotic tasks. For instance, if the robot navigates to the estimated base pose for grasping and discovers that valid Inverse Kinematics (IK) solutions are not available for grasping the object, it necessitates re-planning (updating the object pose and re-estimating the base pose) and re-trying (navigating to the newly estimated base pose). Similarly, a failed grasp not only risks damaging the object but also may disrupt other items in the scene, complicating and delaying the robot's task completion. Hence, in such situations, it is preferable to postpone the execution of actions until there is sufficient confidence regarding the success of the robotic task given the uncertain object pose estimate.

% acceptable error for robotic tasks
The 6D object pose estimation is a challenging problem that depends on 
% several factors such as 
object symmetry, occlusions, lighting conditions, etc. \cite{Xiang2017, tremblay2018deep, shi2021fast}. However, in robotic tasks, a certain level of uncertainty in the object pose estimate is often tolerable \cite{hietanen2021benchmarking, Hagelskjaer2019}. For instance, as shown in Fig.~\ref{fig:idea}b, top row), even with some errors in the grasp pose, the grasp can still succeed. Similarly, despite a certain error in the base pose, Inverse Kinematics solutions may still be available for grasping the object (Fig~\ref{fig:idea}b, bottom row). Evaluating whether the uncertainty in the object pose is acceptable for the success of the robotic task can help prevent task failures or the need for re-planning.

% previous works and limitations
Previous works have attempted to reduce task failures due to uncertain object pose estimates by:
\begin{itemize}[leftmargin=*]
    % re-check first point?
    \item selecting the base or grasp pose (relative to the object) while assuming Gaussian uncertainty in their estimation \cite{kim2012physically, weisz2012pose}.
    \item using one-dimensional predictive confidence provided by the object pose estimators to determine if the action should be executed using the estimated base or grasp pose \cite{shi2021fast}.
\end{itemize}
In practice, uncertainty in the object pose estimate is often multi-modal \cite{deng2021poserbpf, naik2022multi}, and hence the estimated base pose and grasp pose also have multi-modal uncertainties. Similarly, the \textit{acceptable error space} for the robotic task success can also be multi-modal, e.g., grasping the \textit{bowl} object using a top-down grasp (see Fig.~\ref{fig:gecm_illustration}). Further, the predictive confidences provided by the object pose estimators \cite{Xiang2017, shi2021fast, gupta2019cullnet} only considers the confidence in the object pose estimate and does not take into account the error that is acceptable for the success of robotic tasks. For example, the base pose for grasping allows for a larger margin of error compared to the grasp pose as usually there are regions, where many IK solutions exist, whereas grasping is highly sensitive to errors and even minor deviation can lead to a failed grasp. 

% technical motivations for our work
In recent years, significant progress has been made in estimating the full 6D object pose distributions \cite{Manhardt2019, Okorn2020, deng2021poserbpf, murphy2021implicit, naik2022multi, iversen2022ki, haugaard2023spyropose}. These works capture the uncertainty in the object pose by modeling the posterior as a multi-modal distribution. Additionally, significant progress has been made in dynamic simulation tools, enabling reliable predictions of robotic task success in the real world by initially simulating the task \cite{kim2012physically, collins2021review}.

% our contribution
In this work, we present a framework for assessing the likelihood of the robotic task's success by evaluating whether the uncertainty in the estimated object pose is acceptable for successful task completion. We represent both the \textit{estimated error space} of the object pose and the \textit{acceptable error space} for task success using multi-modal non-parametric probability distributions.  Our proposed framework consists of two main steps:
\begin{enumerate}
\item During the offline phase, we analyze the object pose error space and pre-compute the \textit{acceptable error space} (Fig.~\ref{fig:idea}b) for successful task execution using the selected tool pose (Fig.~\ref{fig:idea}a). This involves introducing various errors in the object pose and conducting dynamic simulations to identify acceptable errors for task success.
\item In the online phase, we integrate the \textit{acceptable error space} (Fig.~\ref{fig:idea}b) for successful task execution over the \textit{estimated error space} in the object pose estimate (Fig.~\ref{fig:idea}c), obtained through pose distribution, to predict the likelihood of task success (Fig.~\ref{fig:idea}d).
\end{enumerate}
Through experimental evaluation, we demonstrate that the proposed framework can better predict the success of robotic tasks compared to related works.

% outline
% This paper is structured as follows. In Section~\ref{sec:related_work}, we present related work. We define the problem in Section~\ref{sec:pd}, followed by a description of the proposed approach in Section~\ref{sec:pa}. The experiment setup is presented in Section~\ref{sec:ee}, followed by results in Section~\ref{sec:res}, and conclusions in Section~\ref{sec:con_and_fut}.

\section{RELATED WORK}
\label{sec:related_work}
% In this section, we review the related work on task planning assuming uncertainty in the object pose (Section~\ref{ss:rw:tpapu}), followed by decision-making under pose uncertainty (Section~\ref{ss:rw:dmupu}) and summarize our contributions.

\subsection{Task planning assuming pose uncertainty}
\label{ss:rw:tpapu}
Most related works plan the tool pose (relative to the object) assuming that there will be uncertainty in the estimated tool pose. For instance, in the context of base pose planning, \cite{stulp2012learning, xu2020planning, meng2021uncertainty} assume that there will be uncertainty in the estimated base pose. Consequently, they select a base pose that, despite this uncertainty, still allows for Inverse Kinematics (IK) solutions to grasp the object. All these works assume that the propagated uncertainty in the estimated tool pose follows a Gaussian distribution. In reality, object pose distributions can exhibit multi-modal behavior, which challenges this assumption.

For grasp planning, \cite{kim2012physically, weisz2012pose} have proposed measures for predicting grasp success considering both object dynamics and pose uncertainty. However, similar to the base pose planning works, they rely on the assumption of Gaussian uncertainty. Other works connect errors in object pose and grasp success by modeling pose uncertainty using kernel regression estimated using a real physical setup \cite{hietanen2021benchmarking}.

\noindent
\textbf{Contribution.}
In this work, we do not assume the \textit{acceptable error space} to be a uni-modal Gaussian distribution. Instead, we represent the \textit{acceptable error space} as a multi-modal non-parametric distribution. 
% and use dynamic simulations to assess task success for different errors in the object pose. 
% In this work, we do not assume that only Gaussian uncertainty in the tool pose is acceptable for the success of the robotic task. Instead, we represent the acceptable error space as a multi-modal non-parametric distribution and use dynamic simulations to assess task success for different errors in the object pose. 
% We then integrate the \textit{acceptable error space} over the object pose uncertainty (also represented using a multi-modal distribution) to determine the likelihood of task success.

\subsection{Decision making under pose uncertainty}
\label{ss:rw:dmupu}
Existing works that integrate visual object pose uncertainty into the robot’s decision-making either use one-dimensional predictive confidence obtained using deep learning-based end-to-end pose estimators \cite{Xiang2017, tremblay2018deep}, ensemble methods \cite{shi2021fast}, or by comparison with re-projected CAD models based on the estimated pose \cite{gupta2019cullnet} to predict the success chances of the robotic action. Such predictive confidences only consider confidence in the estimated visual object pose and do not take into account the error that is acceptable for the successful completion of the robotic task. Other works have dealt with uncertain pose estimates using sequential decision-making \cite{naik2024}.

\noindent
\textbf{Contribution.} In this work, instead of relying on confidence measures that lack the capacity to determine if the error is acceptable for the robotic task, we propose a confidence measure that considers the uncertainty in the object pose, represented as a 6D multi-modal non-parametric probability distribution, to assess whether the estimated error is acceptable for the task's success. While \cite{Hagelskjaer2019} simultaneously optimize the pose uncertainty and gripper finger design to accommodate task requirements, to the best of our knowledge, our work is the first to represent both pose uncertainty and acceptable error for the robotic task using multi-modal non-parametric distributions and demonstrate its superior performance on two real-world examples.

% \LN{I think this is similar to illustrative figures so can be excluded!}

% \begin{figure*}[h]
%     \centering
%     \includegraphics[width=17.5cm]{images/proposed-approach/architecture.pdf}
%     \caption{Proposed approach \textbf{a) Error distribution:} Estimating the possible errors $\eest$ in the object pose estimate $\poseestimate{}$ using the pose distribution $\poseuncertainty{}$ \textbf{(b) Task success evaluation:} Determining the acceptable error space $\eacc$ for the task by evaluating the task success offline for different errors in the error space $\errorspace$ \textbf{(c) Probability of task success:} Determining the probability of task success based on the estimated $\eest$ and the acceptable errors $\eacc$.} 
%     \label{fig:architecture}
% \end{figure*}

\section{PROBLEM FORMULATION}
\label{sec:pd}
We address the problem of predicting the success likelihood of a robotic task that requires a 6D pose of the object, given the selected tool pose relative to the object, the estimated object pose $\poseestimate{}$ and the associated uncertainty represented as a probability density function $\poseuncertainty{}$ over SE(3). 

Let $\cam{}$, $\obj{}$, $\taskf{}$, and $\world{}$ be the camera, object, tool, and world frames respectively. The camera frame $\cam{}$ is the frame in which the 6D object pose $\poseestimate{}$ is estimated. The tool frame $\taskf{}$ depends on the robotic task; for example, in the context of determining the base pose for grasping, the tool frame $\taskf{}$ corresponds to the \textit{mobile base} frame $\basef{}$, while for grasping, it corresponds to the \textit{gripper} frame $\gripperf{}$. The world frame $\world{}$ is the frame in which the task is planned and executed; for example, for navigation, the $\world{}$ corresponds to the \textit{map} frame, while for grasping, the $\world{}$ corresponds to the \textit{manipulator base} frame.

To perform the task, the transformation from the world to the tool frame $\Twt$ is required. This transformation can be obtained as follows:
\begin{equation}
\label{eq:Twt}
    \Twt = \Twc \cdot \Tco \cdot \Tot,
\end{equation}
where $\Twc$ is the transformation from the world to the camera frame and is assumed to be known, $\Tco$ is the transformation from the camera to the object and is obtained using the object pose estimate $\poseestimate{}$ and $\Tot$ is the transformation from the object to the tool frame (tool pose relative to the object). For the given task, the tool pose relative to the object $\Tot$ can be determined either manually or using some optimization methods \cite{vahrenkamp2013robot, roa2015grasp}.
% , palleschi2023grasp}. 

% For example, the base pose for grasping can be determined using Inverse Reachability Maps \cite{vahrenkamp2013robot, makhal2018reuleaux} or data-driven methods \cite{jauhri_robot_2022}. Similarly, grasp candidates can be determined either with analytical \cite{rubert2017relevance, roa2015grasp} or data-driven methods \cite{palleschi2023grasp, mousavian20196}. 

As the pose estimate $\poseestimate{}$ is uncertain, the pose estimate $\poseestimate{}$ represents the transformation from the camera frame $\cam{}$ to the estimated object frame $\obje{}$
\begin{equation}
    \poseestimate{} = \Tcoe.
\end{equation}
% (and not $\Tco$). 
Further, any sample $\samplecont \in SE(3)$ from the pose distribution $\poseuncertainty{}$ could be the true transformation from the camera to the object $\Tco$. Thus, 
\begin{equation}
    \samplecont = \Tco = \Tcoe \cdot \Toeo,
\end{equation}
where $\Toeo=\error$ is the estimated error in the object pose estimate $\poseestimate{}$. 
% \clarify{(assuming $\samplecont$ to be the true object pose)}. 
This error can result in incorrect $\Twt$ (Eq.~\ref{eq:Twt}).
% \begin{equation}
%     \Twt = \Twc \cdot \Tcop \cdot \Topt,
% \end{equation}
% where $\Topt = \Topo \cdot \Tot$. 
All the samples $\samplecont$ in $\poseuncertainty{}$ represent possible errors $\error$ in the pose estimate $\poseestimate{}$ (\textit{estimated error space}). Thus, the probability of task success using the estimated object frame $\obje{}$, $\prob(\task|\obje{})$, can be calculated by marginalizing over 
% all the possible errors $\error$ modeled by the pose distribution $\poseuncertainty{}$
the \textit{estimated error space}
\begin{equation}
\begin{aligned}
    \label{eq:task_success_cont}
    \prob(\task|\obje{}) = & \int \prob(\task|\obje{}, \Toeo) \cdot \prob(\Toeo|\obje{}) \; d\Toeo \\
                     = & \int \prob(\task|\obje{}, \error) \cdot \prob(\error|\obje{}) \; d\error,
\end{aligned}
\end{equation}
% \begin{equation}
% \begin{aligned}
%     \label{eq:task_success_cont}
%     \prob(\task|\obje{}) = \int \prob(\task|\obj{}, \error) \cdot \prob(\error|\obje{}) \; d\error,
% \end{aligned}
% \end{equation}
where $\prob(\task|\obje{}, \error)$ is the probability of task success given the error $\error$ in the estimated object frame $\obje{}$ (\textit{acceptable error space})
% \footnote{In $\prob(\task|\obj{}, \error)$, $\obj{}$ refers to the ground truth object frame.}
, and $\prob(\error|\obje{})$ is the probability of error $\error$ in the estimated object frame $\obje{}$, calculated using the pose distribution $\poseuncertainty{}$.

% The task success given the error $\error$ can be evaluated in simulation by introducing the error $\error$ in the tool pose $\Tot$ and then performing the task using the perturbed tool pose 
% % $\Topt= 
% ($\error^{-1} \cdot \Tot = \Tooe \cdot \Tot$). However, performing dynamic simulations online is computationally expensive as pose distributions $\poseuncertainty{}$ can contain millions of samples \cite{deng2021poserbpf}. 

\section{PROPOSED APPROACH}
\label{sec:pa}
The task success given the error $\error$ in the pose estimate $\poseestimate{}$ can be evaluated in simulation by propagating the error $\error$ to the tool pose and then performing the task using the perturbed tool pose. However, performing dynamic simulations online for all possible errors is computationally expensive as pose distributions $\poseuncertainty{}$ can contain millions of samples \cite{deng2021poserbpf}. We address this by performing dynamic simulations offline to create a mapping from error space to the task success likelihood and then only integrate these errors over the estimated pose uncertainty $\poseuncertainty{}$ online to determine the task success likelihood using the pose estimate $\poseestimate{}$. 

We consider the error space of error $\error$ in the object frame $\obj{}$ to be discrete. This allows $\poseuncertainty{}$ to be multi-modal. Let set $\errorspace$ represent the discrete error space, such that $|\errorspace|=N$ and each element $\errorn \in SE(3)$ is a transformation from the perturbed object frame $\objn{}$ to the true object frame $\obj{}$, $\Tono$, represented as:
\begin{equation}
\errorn = \{ \ecomp{t}{x}, \ecomp{t}{y}, \ecomp{t}{z}, \ecomp{r}{x}, \ecomp{r}{y}, \ecomp{r}{z} \},
\end{equation}
where $\ecomp{t}{x} \in [-\ein{t}{x}..\ein{t}{x}]$, $\ecomp{t}{y} \in [-\ein{t}{y}..\ein{t}{y}]$, $\ecomp{t}{z} \in [-\ein{t}{z}..\ein{t}{z}]$, $\ecomp{r}{x} \in [-\ein{r}{x}..\ein{r}{x}]$, $\ecomp{r}{y} \in [-\ein{r}{y}..\ein{r}{y}]$, and $\ecomp{r}{z} \in [-\ein{r}{z}..\ein{r}{z}]$. The translation error $(\ecomp{t}{x}, \ecomp{t}{y}, \ecomp{t}{z})$ is sampled with a discretization factor of $\tdf$, and the rotational error $(\ecomp{r}{x}, \ecomp{r}{y}, \ecomp{r}{z})$ (represented as Euler angles) is sampled with a discretization factor of $\rdf$. 

This error space is used for computing the \textit{acceptable error space} (Section~\ref{ss:pa:etss}) and the \textit{estimated error space} (Section~\ref{ss:pa:pd}). The probability of task success is then calculated by integrating the \textit{acceptable error space} over the \textit{estimated error space} (Section~\ref{ss:pa:pts}).

% This error space is used to compute the \textit{acceptable error space} for the success of the robotic task (Section~\ref{ss:pa:etss}) and the \textit{estimated error space} for the object pose estimate $\poseestimate{}$ using the pose distribution $\poseuncertainty{}$ (Section~\ref{ss:pa:pd}). Finally, we integrate the \textit{acceptable error space} over the \textit{estimated error space} to calculate the probability of task success (Section~\ref{ss:pa:pts}).

% The proposed approach consists of three steps. First, we obtain the \textit{estimated error space} for the object pose estimate $\poseestimate{}$ using the pose distribution $\poseuncertainty{}$ (Section~\ref{ss:pa:pd}). Next, we use dynamic simulations to analyze the \textit{acceptable error space} for the success of the robotic task (Section~\ref{ss:pa:etss}). Finally, we map the \textit{estimated error space} to the \textit{acceptable error space} to calculate the probability of task success (Section~\ref{ss:pa:pts}).

\subsection{Acceptable error space}
\label{ss:pa:etss}

% All possible errors $\error$ in the object pose estimate $\poseestimate{}$, modeled by the pose distribution $\poseuncertainty{}$, could be in principle introduced into the simulator, to task success given this error $\error$ can be evaluated \cite{kim2012physically, weisz2012pose}. However, this becomes computationally intractable when considering pose distributions with millions of samples \cite{deng2021poserbpf}. Our discrete error space $\errorspace$ allows us to perform these evaluations offline.

This is an offline step, wherein we pre-determine the errors $\errorn \in \errorspace$ in the object pose that are acceptable for the robotic task i.e. despite the error $\errorn$ in object pose, the task can be successfully performed.

We approximate the probability of task success $\prob(\task|\obje{}, \error)$ by the probability of the closest discrete error $\errorn$:
\begin{equation}
\prob(\task|\obje{}, \error) \approx \prob(\task|\obje{}, \errorn),
\end{equation}
where 
% $\prob(\task|\objp{}, \errorn)$ is the probability of task success given error $\errorn$ in the object frame $\objp{}$ and 
$\errorn$ is selected as:
\begin{equation*}
\errorn = \argmin_{\errorn \in \errorspace}\; \mathbf{D}(\error,\errorn).
\end{equation*}
The task success given the error $\errorn$ in the object frame estimate $\obje{}$ is evaluated in simulation by calculating the tool pose relative to the perturbed object frame $\objn{}$ ($\Toon \cdot \Tot$) and then performing the task using the perturbed tool pose. 
% introducing the error $\errorn$ in the tool pose $\Tot$ and then performing the task using the perturbed tool pose 
% % $\Topt= 
% ($\error^{-1} \cdot \Tot = \Tooe \cdot \Tot$).
When evaluated in simulation, this results either in task success or failure: 
\begin{equation}
\prob(\task|\obje{}, \errorn): \mathrm{SE}(3) \rightarrow \{0, 1\}.
\end{equation}
Let $\eacc \subset \errorspace$ be the \textit{acceptable error space} containing errors $\errorn \in \errorspace$ that are acceptable for the success of the robotic task. Then $\eacc$ is defined as:
\begin{equation}
\eacc = \{ \errorn \in \errorspace | \prob(\task|\obje{}, \errorn) = 1 \}. % Should these be 2 equal signs?
\end{equation}
$\prob(\task|\obje{}, \errorn)$ can be represented as an indicator function:
\begin{equation}
\prob(\task|\obje{}, \errorn) = \textbf{1}_{\eacc}(\errorn),
\end{equation}
which returns $True$ if $\errorn \in \eacc$.

\subsection{Estimated error space}
\label{ss:pa:pd}
This is an online step, wherein we determine the possible errors $\error$ in the estimated pose $\poseestimate{}$ using the pose distribution $\poseuncertainty{}$ which is estimated using \cite{deng2021poserbpf, naik2022multi}.

The error distribution $\prob(\errorn|\obje{})$ provides the probability of error $\errorn \in \errorspace$ given the estimated object frame $\obje{}$. 
% given the pose estimate $\poseestimate{}$ and distribution $\poseuncertainty{}$. 
% The possible errors $\error$ in the object pose are estimated by comparing the estimated object pose $\poseestimate{}$ ($\Tcop$) with pose samples $\samplecont(\Tco$) in the full pose distribution $\poseuncertainty{}$. The error distribution $\prob(\errorn)$ 
It is obtained as:
\begin{equation}
\prob(\errorn|\obje{}) = \prob(X=\argmin_{\samplecont \in X}\; \mathbf{D}(\poseestimate{}^{-1} \cdot \samplecont,\errorn)),
\end{equation}
where $\poseestimate{}=\Tcoe$, $\samplecont=\Tco$, $\errorn=\Tono$ and $\mathbf{D}(a,b)$ is the distance metric between $a$ and $b$ in $SE(3)$. Finally, the \textit{estimated error space} $\eest \subset \errorspace$ is defined as,
\begin{equation}
\eest = \{ \errorn \in \errorspace | \prob(\errorn|\obje{}) > p_{\text{thres}} \},
\end{equation}
where $p_{\text{thres}}$ is a threshold for discarding the samples with very small probabilities.

% \clarify{\textbf{Doubt}: Up to this point, $\obj{}$ is a variable that can assume any possible (true) object poses modeled by the pose distribution. However, in the next section $\obj{}$ is a constant and represents only the actual true object pose (ground truth). Is this okay or can create confusion?}

\subsection{Probability of task success under pose uncertainty}
\label{ss:pa:pts}
This is the online step, wherein the probability of task success under pose uncertainty is calculated using Eq.~\ref{eq:task_success_cont}. Upon discretizing the error space, Eq.~\ref{eq:task_success_cont} can be reformulated as:
\begin{equation}
\begin{aligned}
    \label{eq:task_success_disc}
\prob(\task|\obje{}) = & \sum_{\errorspace} \prob(\task|\obje{}, \errorn) \cdot \prob(\errorn|\obje{}) \\
                     = & \sum_{\errorn \in \eest}^{} \textbf{1}_{\eacc}(\errorn) \cdot \prob(\errorn|\obje{})
\end{aligned}
\end{equation}
where $\textbf{1}_{\eacc}(\errorn)$ indicates if the error is acceptable for the task (\textit{acceptable error space} in Section~\ref{ss:pa:etss}) and $\prob(\errorn|\obje{})$ is obtained using the pose estimate $\poseestimate{}$ and distribution $\poseuncertainty{}$ (\textit{estimated error space} in Section~\ref{ss:pa:pd}). Essentially, Eq.~\ref{eq:task_success_disc} integrates the pre-computed \textit{acceptable error space} $\eacc$ over the \textit{estimated error space} $\eest$ in the object pose to determine the probability of task success.

\section{Applicability to Different Tasks}
\label{sec:illus}
In this Section, we demonstrate the proposed generic framework on two illustrative examples for the tasks discussed in Section~\ref{sec:intro} (Fig.~\ref{fig:idea}).
% : the \textit{availability of IK solution} at the estimated base pose for grasping and 2) the \textit{grasp success} using the estimated grasp pose. 

\subsection{Availability of IK solution}
To be able to grasp the object from the estimated base pose, there must be a valid IK solution to reach the grasp pose. As the base pose is defined relative to the object $\Tob$ and estimated using the object pose estimate $\poseestimate{}$ (which provides object frame estimate $\obje{}$), any error $\error$ in $\poseestimate{}$ is also propagated to the estimated base pose. 

\begin{figure}[h]
    \centering
    \includegraphics[width=8.5cm]{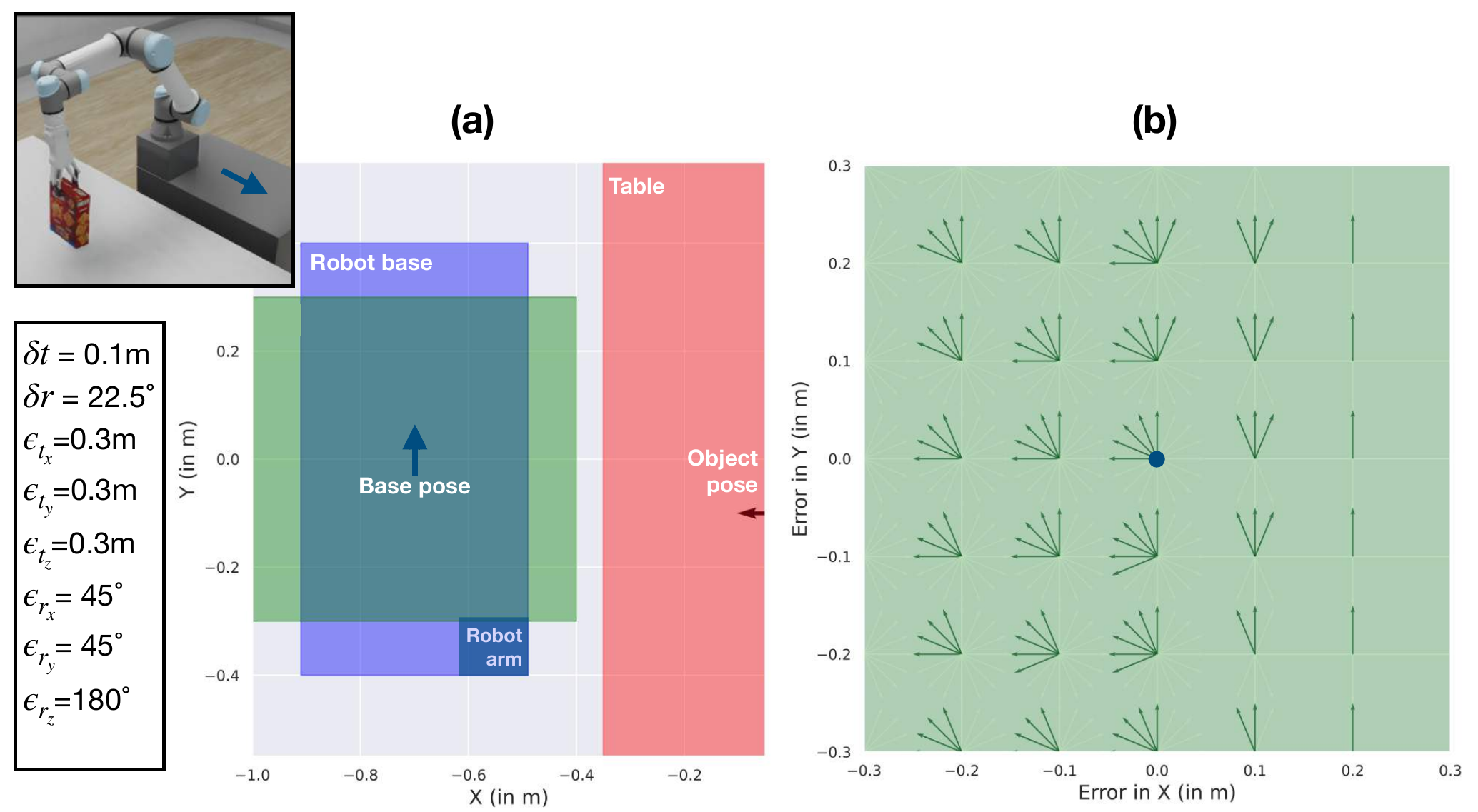}
    \caption{\textbf{a)} Selected base pose $\Tob$ and top-down view of the grasping scene. Error space $\errorspace$ (green box) is defined around the selected base pose for grasping (blue arrow at the center of robot base) \textbf{b)} \textit{Acceptable error space} $\eacc$.}
    \label{fig:irm_illustration}
\end{figure}

The availability of IK solution at the selected base pose relative to the object $\Tob$ for different errors $\errorn$ in the object frame can be evaluated using kinematics solver \cite{aristidou2011fabrik}. This can be used for pre-computing the \textit{acceptable error space} $\eacc$.
% Task failures due to errors $\error$ can be prevented by pre-computing the acceptable error space $\eacc$ (Fig.~\ref{fig:irm_illustration}b) for the selected base pose for grasping $\Tob$ (Fig.~\ref{fig:irm_illustration}a). 
% In Fig.~\ref{fig:irm_illustration}, we illustrate the acceptable error space for the base pose for the grasping task. 
%Discretization factors $\tdf$ and $\rdf$ were set to 10 cm and $\pi$/8 respectively. Translational limits ($\ecomp{t}{x},\ecomp{t}{y},\ecomp{t}{z}$) were set to 30cm, and rotational limits ($\ecomp{r}{x},\ecomp{r}{y},\ecomp{r}{z}$) were set to $\pi$. 
Fig.~\ref{fig:irm_illustration}b describes the \textit{acceptable error space} $\eacc$ for the base pose in Fig.~\ref{fig:irm_illustration}a. The majority of base poses in the $\eacc$ (Fig.~\ref{fig:irm_illustration}b) are concentrated in the central region. On the right, most base poses lead to collisions with the table, while on the left, reachability diminishes as the distance to the object increases.
% The $\eacc$ (Fig.~\ref{fig:irm_illustration}b) has very few valid base poses to the right as most base poses will result in the robot colliding with the table. The majority of valid base poses are concentrated in the central area. Moving from the central region towards the left, the number of valid base poses decreases due to the increasing distance between the object and the robot, resulting in a decrease in the robot's reachability.

The probability of an IK solution using the object frame estimate $\prob(\text{IK}|\obje{})$ can then be calculated by mapping the \textit{estimated error space} $\eest$ onto the \textit{acceptable error space} $\eacc$ as shown in Fig.~\ref{fig:irm_illustration}.
\begin{figure}[h]
    \centering
    \includegraphics[width=8.5cm]{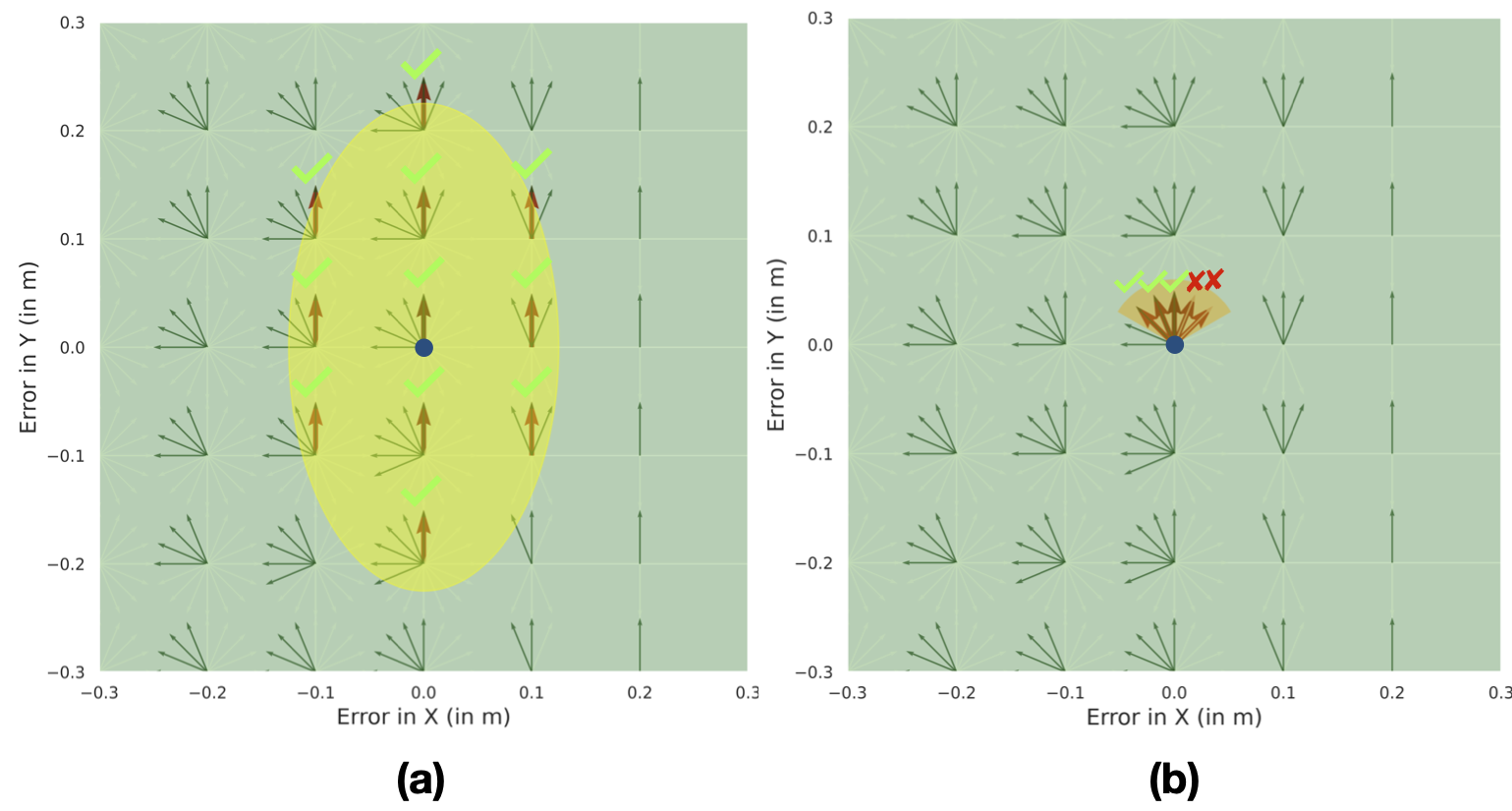}
    \caption{Calculating the $\prob(\text{IK}|\obje{})$ \textbf{a)} \textit{Estimated error space} $\eest$ (yellow ellipsoid) mapped onto the \textit{acceptable error space} $\eacc$. \textbf{b)} \textit{Estimated error space} $\eest$ (orange cone) mapped onto the \textit{acceptable error space} $\eacc$.}
    \label{fig:irm_success}
\end{figure}
For example, if $\eest$ only exhibits significant translation uncertainty (depicted by the yellow ellipsoid in Fig.~\ref{fig:irm_success}a) but no uncertainty in rotation, $\prob(\text{IK}|\obje{})$ will be 1.0 as all potential base poses (represented in red) yield valid IK solutions. However, if $\eest$ only exhibits rotational uncertainty of approximately 45$^{\circ}$ with no translation uncertainty (depicted by an orange cone in Fig.~\ref{fig:irm_success}b), then an IK solution is only possible for three out of the five potential base poses, i.e. $\prob(\text{IK}|\obje{})$=0.6.

\subsection{Grasp success}
For a successful grasp, the gripper fingers must enclose the object with stable contact points \cite{kim2012physically, weisz2012pose}. Similar to the base pose, the grasp pose is also defined relative to the object $\Tog$ and estimated using the object pose estimate $\poseestimate{}$. Consequently, any error $\error$ in $\poseestimate{}$ is also propagated to the estimated grasp pose.

\begin{figure}[h]
    \centering
    \includegraphics[width=8.6cm]{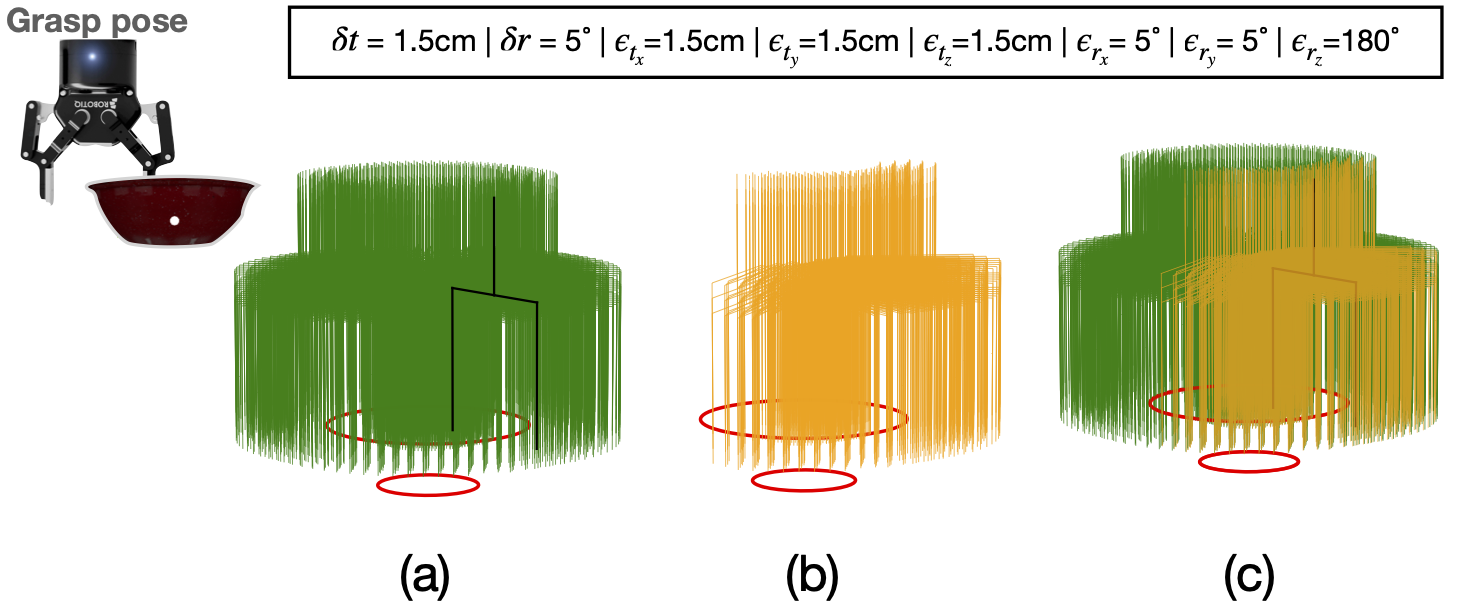}
    \caption{Calculating the probability of grasp success for the \textit{bowl} object. \textbf{a)} Selected grasp pose and \textit{Acceptable error space} $\eacc$ \textbf{b)} \textit{Estimated error space} $\eest$ \textbf{c)} $\eacc$ (green) integrated over $\eest$ (orange) for calculating $\prob(\text{grasp}|\obje{})$.}
    \label{fig:gecm_illustration}
\end{figure}

The grasp success using the selected grasp pose $\Tog$ for different errors $\errorn$ in the object can be evaluated using dynamic simulation to pre-compute \textit{acceptable error space} $\eacc$ (Fig.~\ref{fig:gecm_illustration}a). In Fig.~\ref{fig:gecm_illustration}a, the red circles denote the top and bottom of the bowl object, the black gripper symbol denotes the selected grasp pose $\Tog$, and each green gripper symbol is a grasp pose relative to perturbed object frames $\objn{}$ (due to errors $\errorn$ in the object frame $\obj{}$) that would result in a successful grasp.

The probability of a grasp success using the estimated object frame $\prob(\text{grasp}|\obje{})$ can then be calculated by integrating the \textit{acceptable error space} $\eacc$ over the \textit{estimated error space} $\eest$ (Fig.~\ref{fig:gecm_illustration}b) as shown in Fig.~\ref{fig:gecm_illustration}c. In Fig.~\ref{fig:gecm_illustration}c, all of the resultant grasp poses result in a successful grasp, thus $\prob(\text{grasp}|\obje{})$ is 1.00.

\section{EXPERIMENTAL EVALUATION}
\label{sec:ee}

\subsection{Experiment Objectives and Baselines}
\label{ss::ee:eob}
Our experiments are designed to demonstrate that:
\begin{itemize}[leftmargin=*]
\setlength\itemsep{0.25em}
\item Our framework can better predict task success under pose uncertainty as it considers both the uncertainty in the object pose and the acceptable errors for task success. 
\item Representing the acceptable error space for task success using a multi-modal distribution leads to better decisions, as it can better capture the task success requirements.
% more task attempts and fewer views for decision-making as it can better capture the acceptable error space compared to uni-modal distributions.
% \item Assuming that Gaussian uncertainty in the object pose is acceptable for robotic task success is unreliable as:
% \begin{enumerate}
% \item It can lead to unnecessary delays in task execution, as multi-modal pose distributions will only converge to a uni-modal distribution when unique views of the object become available (assuming the object has unique views).
% \item It can lead to failures if multi-modal distributions falsely collapse to a uni-modal distribution, for example, in the case of occlusions.
% \item It may never attempt the task for objects like mugs, which lack a unique view, and hence ideally distributions will always be multi-modal.
% \end{enumerate}
\end{itemize}
We consider three baselines:
\begin{itemize}[leftmargin=*]
    \setlength\itemsep{0.25em}
    \item[] \textbf{Blind execution (BE):} The robot blindly executes the task using the estimated object pose $\poseestimate{}$.
    \item[] \textbf{Visual confidence (VC):} The robot uses the confidence provided by the pose estimator, which depends on the extent of object occlusion \cite{deng2021poserbpf} to decide if the task should be executed using the estimated object pose $\poseestimate{}$. 
    \item[] \textbf{Gaussian uncertainty (GU):} The robot assumes that the \textit{acceptable error space} can be represented by a uni-modal Gaussian distribution. Therefore, it executes the task only if the estimated uncertainty in the object pose is also uni-modal. The nature of the pose distribution is determined using the normality test \cite{henze1990class}.
\end{itemize}

\subsection{Evaluation metrics}
\label{ss::ee:eob}
We use the following metrics for evaluation:
\begin{itemize}%[leftmargin=*]
    \setlength\itemsep{0.25em}
    \item \textit{Attempts:} number of times the decision was made to execute the task.
    \item \textit{Successes:} number of times the task was successfully executed after the decision to execute the task.
    \item \textit{Failures:} number of times the task failed after the decision to execute the task.
    \item \textit{Avg. views:} average number of views used to decide on the task execution
\end{itemize}
We demonstrate the first objective by showing that our method results in larger \textit{successes} and fewer \textit{failures} compared to the baselines. It is difficult to validate the second objective quantitatively as the decisions depend on the context. Hence, we use specific examples using the evaluation metrics defined above. Additionally, we present two qualitative examples.

% However, as many times pose distribution falsely collapses to the uni-modal distribution, leading to early decisions (and sometimes failures), avg. number of views is not a completely reliable metric. Hence, we also present 2 qualitative examples to support this claim.

% To show that there is a better correlation between task success and our score compared to standard metrics used for measuring pose estimation accuracy. (here we can also do one where uncertainty was correctly modeled and not)

\subsection{Experiment Setup}
\label{ss::ee:es}
We used two different experiment setups for both tasks. For the availability of the IK solution task, we assume that the robot chooses to estimate the base pose at a distance of approximately 3m from the object (Fig.~\ref{fig:base_qual}). If the decision cannot be made, it progressively approaches the object until the decision can be made. To prevent the influence of errors due to robot localization inaccuracies and the criteria used by the navigation stack to determine if the goal is reached, we chose a hybrid setup for evaluation. While these are valid challenges in mobile robotics, in this study we only consider uncertainty in the object pose estimate $\poseestimate{}$. 

We used the MY-MM dataset\footnote{\url{https://lakshadeep.github.io/mymm/}}, which provides ground truth data for object pose tracking from a distance through the robot's onboard camera and external cameras in the environment. Due to the larger distance between the object and the robot's onboard cameras, we used the robot's onboard camera and external cameras for estimating the object pose and distribution using \cite{naik2022multi}. We assessed the availability of IK solution at the estimated base pose using the ground truth object poses provided by the dataset (see Fig.~\ref{fig:base_qual}). 

% The object pose and its distribution were estimated using multi-view object pose distribution tracking \cite{naik2022multi}. When there is insufficient confidence to execute the task, the decision is postponed to the subsequent frame, and so forth. Ground truths were employed to replicate the scene in NVIDIA Isaac sim \footnote{\url{https://developer.nvidia.com/isaac-sim}}. Upon deciding to navigate to the estimated base pose, the robot was teleported to this base pose, and we verified if there was an IK solution to reach the grasp pose computed using the ground truth object pose (See Fig.~\ref{fig:base_qual}).

\begin{figure*}[h]
    \centering
    \includegraphics[width=17.5cm]{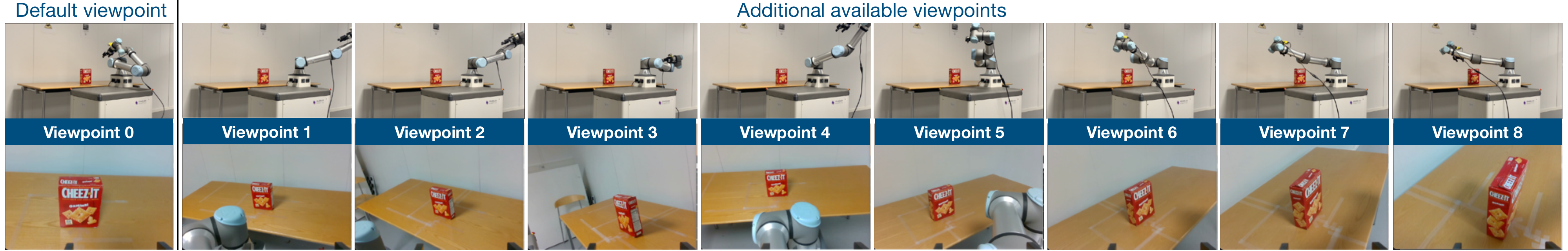}
    \caption{\textbf{Setup for grasping experiments:} The default viewpoint and additional available viewpoints. The next viewpoint is selected according to the above order if the robot decides not to perform the task using the pose estimate $\poseestimate{}$ and distribution $\poseuncertainty{}$ estimated from the current viewpoint.}
    \label{fig:grasp_setup}
\end{figure*}

For the grasp success task, we assume that the robot has already reached the base pose for grasping. As the robot camera is close to the object, we estimated the pose and distribution only with the robot camera using \cite{deng2021poserbpf}. We defined eight distinct camera viewpoints, each offering a unique view of the object as shown in Fig.~\ref{fig:grasp_setup}. For each object, 20 trials were conducted: in 4 trials, there was no occlusion in the default viewpoint, in 8 trials, around 30 percent of the object was occluded, and in 8 trials, around 50 percent of the object was occluded. 

For both tasks, the `BE' performs the task using the object pose estimated in the default viewpoint (unless it is not possible to execute the action, for example, if the estimated base pose will result in the robot colliding with the table). For our method (`Ours') and the other two baselines (`VC' and `GU'), the robot transitions to the next viewpoint if the measured confidence is below the pre-defined threshold, and repeats the process. If none of the viewpoints yields sufficient confidence, the task is not performed.

% It should be noted that in the MY-MM dataset, the object always remains within the field of view of the camera as the robot approaches it. Consequently, continuous pose and distribution tracking were achievable. However, during grasp experiments, it was not always feasible to guarantee that the object remained within the field of view, as the views were highly distinct. Thus, tracking was often re-initialized after transitioning to a new viewpoint.

% For the cracker box object (Fig.~\ref{fig:gac_t_0}f), there is a high likelihood of grasp success in the front and back of the sphere. This is due to successful grasping occurring only when the gripper aligns with the x-axis of the object with some tolerance, while failure is likely if the rotational error around the z-axis increases. Observing the bowl and mug (Fig.\ref{fig:gac_t_0}f and Fig.\ref{fig:gac_t_0}g), it can be noted that they can tolerate a significant amount of uncertainty in rotational error around the z and y axes, except near the handle in the case of the mug.
% % Since the gripper opening is much larger than the handle of the mug, it doesn't pose any problem for grasping. 
% In contrast, for the ``mustard bottle" and ``bleach cleanser" objects, the gripper's ability to compensate for errors is quite limited, as evident in Fig.~\ref{fig:gac_t_0}g and h. Such maps were pre-computed for different combinations of translational errors across the three axes.

\subsection{Implementation details}
\label{ss::ee:id}

\begin{figure}[h]
    \centering
    \includegraphics[width=8.5cm]{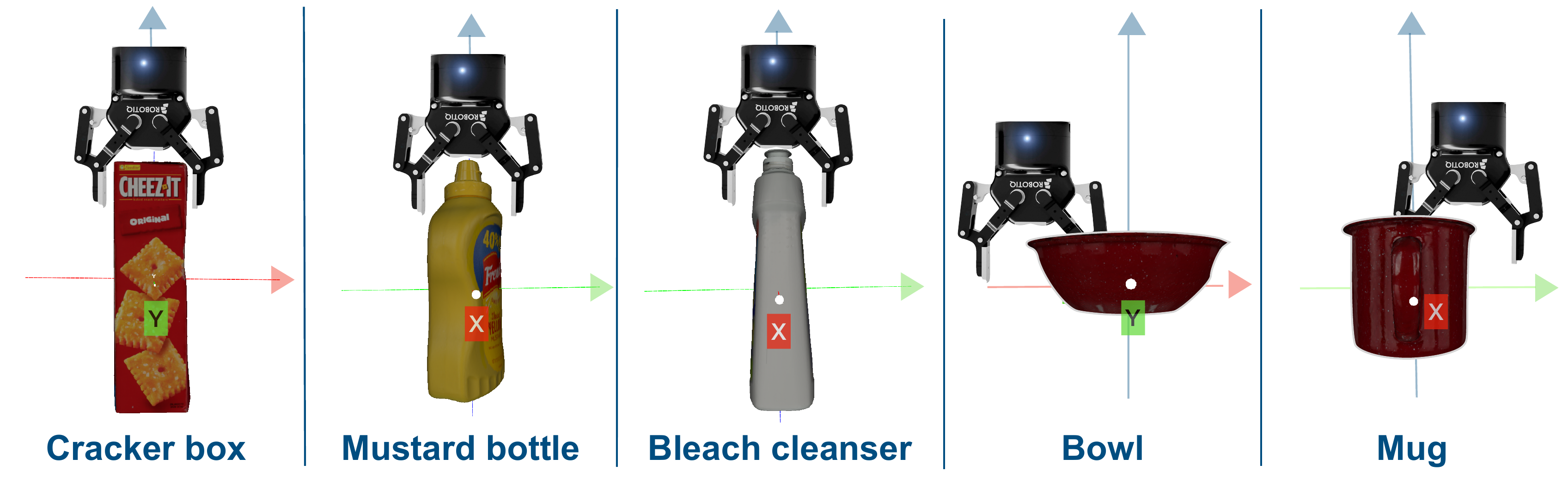}
    \caption{Selected header (top-down) grasp poses $\Tog$ for different objects.}
    \label{fig:gac_t_0}
\end{figure}

% We validate our method and all the baselines on both tasks discussed in Section~\ref{sec:illus}. 
We used 5 different objects from YCB benchmark \cite{calli2015benchmarking} (`cracker box', `mustard bottle', `bleach cleanser', `bowl', and `mug') for evaluation.  For all the objects, we used the header (top-down) grasp pose as shown in Fig.~\ref{fig:gac_t_0}. 

% \footnote{\url{https://developer.nvidia.com/isaac-sim}}
The \textit{acceptable error space} $\eacc$ for the grasp success task was computed using the NVIDIA Isaac Sim simulator\footnote{As the dynamic model of the Robotiq 2F gripper (used in real-world experiments) was not available, OnRobot 2F gripper was used for computing the \textit{acceptable error space} for the grasp success. Both grippers have similar finger sizes and the opening distance was fixed.}, while $\eacc$ for the availability of IK solution task was computed using the Lula Kinematics Solver available in NVIDIA Isaac Sim. The same discretization factors and limits were used for the availability of the IK solution task as described in Fig.~\ref{fig:irm_illustration}. For the grasp success task, $\ecomp{r}{x}$ and $\ecomp{r}{y}$ in Fig.~\ref{fig:gecm_illustration} were set to 60$^\circ$. All other errors were deemed unacceptable.

% both the tasks as described in Fig.~\ref{fig:irm_illustration} and \ref{fig:gecm_illustration}.

% The availability of IK solution was evaluated the errors where the resultant base pose didn't result in a collision with static objects in the scene such as a table and the robot base rests on the surface. The same discretization factors and limits were used for both tasks as described in Section~\ref{sec:illus}. 

In an ideal scenario, the robot should decide to execute the task when the probability of task success using the estimated object pose $\poseestimate{}$ reaches 1.0. However, in practice, this is not possible as the object pose distributions are noisy. For objects such as the `cracker box' (large size and distinct texture on all sides), more accurate pose distributions can be obtained compared to objects like `mug' or `bowl' (metallic surface, smaller size, and no unique texture). Hence, when the object was fully visible and a good pose estimate was available, for the `cracker box' success probability consistently reached above 90\% while for `mug' and `bowl' it only rarely reached above 70\%. Hence the success probability threshold was set to 0.6 for all the objects and the task was executed if the probability exceeded this threshold.

\section{RESULTS}
\label{sec:res}

\subsection{Quantitative Results}
In Table~\ref{tab:grasp-results}, we present quantitative results for both tasks across all three baselines and our method for all five objects. For the grasp success task, `Ours' significantly outperforms all three baselines, achieving 91 successful grasps with only 6 failures in 100 trials. This shows that our method can better predict the success of the tasks compared to the baselines. For the availability of the IK solution task,  `Ours' performs slightly better compared to the baselines, with 48 successes and no failures in 50 trials. The baselines perform better (compared to the grasp success task) because the acceptable error for this task is quite high. 
% Moreover, the robot decides to navigate to the estimated base pose only if it will not result in a collision with the table; otherwise, it postpones the decision\footnote{Explaining why 'Blind execution' requires an average of 19.0 frames for decision-making instead of just 1 frame}. 

\begin{table*}[h]
\centering
\resizebox{0.85\textwidth}{!}{%
\begin{tabular}{@{}l|l|
>{\columncolor[HTML]{FFFFFF}}c 
>{\columncolor[HTML]{FFFFFF}}c 
>{\columncolor[HTML]{FFFFFF}}c 
>{\columncolor[HTML]{FFFFFF}}c 
>{\columncolor[HTML]{FFFFFF}}c |ccccc@{}}
\toprule
 &
   &
  \multicolumn{5}{c|}{\cellcolor[HTML]{FFFFFF}\textbf{Successful grasp using the estimated grasp pose}} &
  \multicolumn{5}{c}{\textbf{Availability of IK solution at the estimated base pose}} \\ \cmidrule(l){3-12} 
\multirow{-2}{*}{\textbf{Method}} &
  \multirow{-2}{*}{\textbf{Object}} &
  \textit{Trials} &
  \textit{Attempts} &
  \textit{Successes} &
  \textit{Failures} &
  \textit{Avg. views} &
  \textit{Trials} &
  \textit{Attempts} &
  \textit{Successes} &
  \textit{Failures} &
  \textit{Avg. views} \\ \midrule
 & Cracker box     & 20 & 20 & 14 (1)  & 6  & 1    & 10 & 10 & 9  & 1 & 3.2   \\
 & Mustard bottle  & 20 & 20 & 5(0)    & 15 & 1    & 10 & 10 & 9  & 1 & 8.3   \\
 & Bleach cleanser & 20 & 20 & 14 (7)  & 6  & 1    & 10 & 10 & 8  & 2 & 12.8  \\
 & Bowl            & 20 & 20 & 8(0)    & 12 & 1    & 10 & 9  & 9  & 0 & 53.5  \\
 & Mug             & 20 & 20 & 17 (6)  & 3  & 1    & 10 & 10 & 10 & 0 & 17.4  \\ \cmidrule(l){2-12} 
\multirow{-6}{*}{\textit{\begin{tabular}[c]{@{}l@{}}Blind \\ Execution\\ (BE)\end{tabular}}} &
  \textit{All} &
  \textit{100} &
  \textit{100} &
  \textit{58 (14)} &
  \textit{42} &
  \textit{1} &
  \textit{50} &
  \textit{49} &
  \textit{45} &
  \textit{4} &
  \textit{19.0} \\ \midrule
 & Cracker box     & 20 & 20 & 15 (1)  & 5  & 1.7  & 10 & 10 & 9  & 1 & 3.4   \\
 & Mustard bottle  & 20 & 20 & 12 (3)  & 8  & 1.7  & 10 & 10 & 9  & 1 & 8.3   \\
 & Bleach cleanser & 20 & 20 & 16 (11) & 4  & 1.3  & 10 & 10 & 8  & 2 & 12.8  \\
 & Bowl            & 20 & 20 & 9(0)    & 11 & 1    & 10 & 9  & 9  & 0 & 53.5  \\
 & Mug             & 20 & 20 & 20 (9)  & 0  & 1.4  & 10 & 10 & 10 & 0 & 18.0  \\ \cmidrule(l){2-12} 
\multirow{-6}{*}{\textit{\begin{tabular}[c]{@{}l@{}}Visual\\ Confidence\\ (VC)\end{tabular}}} &
  \textit{All} &
  \textit{100} &
  \textit{100} &
  \textit{72 (24)} &
  \textit{28} &
  \textit{1.42} &
  \textit{50} &
  \textit{49} &
  \textit{45} &
  \textit{4} &
  \textit{19.2} \\ \midrule
 & Cracker box     & 20 & 20 & 18 (2)  & 2  & 2.5  & 10 & 10 & 10 & 0 & 20.1  \\
 & Mustard bottle  & 20 & 18 & 13(0)   & 5  & 5.8  & 10 & 10 & 10 & 0 & 46.8  \\
 & Bleach cleanser & 20 & 20 & 16 (9)  & 4  & 1.9  & 10 & 10 & 7  & 3 & 63.7  \\
 & Bowl            & 20 & 13 & 9(0)    & 4  & 7.1  & 10 & 7  & 7  & 0 & 144.0 \\
 & Mug             & 20 & 20 & 19 (10) & 1  & 1.15 & 10 & 10 & 10 & 0 & 90.1  \\ \cmidrule(l){2-12} 
\multirow{-6}{*}{\textit{\begin{tabular}[c]{@{}l@{}}Gaussian\\ Uncertainty\\ (GU)\end{tabular}}} &
  \textit{All} &
  \textit{100} &
  \textit{91} &
  \textit{75 (21)} &
  \textit{16} &
  \textit{3.69} &
  \textit{50} &
  \textit{47} &
  \textit{44} &
  \textit{3} &
  \textit{72.9} \\ \midrule
 & Cracker box     & 20 & 20 & 19 (3)  & 1  & 1.75 & 10 & 10 & 10 & 0 & 87.0  \\
 & Mustard bottle  & 20 & 17 & 15 (4)  & 2  & 5.2  & 10 & 10 & 10 & 0 & 28.0  \\
 & Bleach cleanser & 20 & 20 & 20 (5)  & 0  & 3.35 & 10 & 10 & 8  & 0 & 72.1  \\
 & Bowl            & 20 & 20 & 17(0)   & 3  & 2.55 & 10 & 8  & 8  & 0 & 148.3 \\
 & Mug             & 20 & 20 & 20 (7)  & 0  & 1.75 & 10 & 10 & 10 & 0 & 74.5  \\ \cmidrule(l){2-12} 
\multirow{-6}{*}{\textit{Ours}} &
  \textit{All} &
  \textit{100} &
  \textit{97} &
  \textit{\textbf{91 (19)}} &
  \textit{\textbf{6}} &
  \textit{2.92} &
  \textit{50} &
  \textit{48} &
  \textit{\textbf{48}} &
  \textit{\textbf{0}} &
  \textit{81.97} \\ \bottomrule
\end{tabular}%
}
\caption{Task success evaluation. () indicates the no. of successful unstable grasps.}
\label{tab:grasp-results}
\end{table*}

Further, our method has higher attempts for both tasks than the `GU' baseline. Specifically for the `bowl' object, the `GU' baseline attempts grasp only 13 times. This is because when the pose distribution is correctly estimated, it is multi-modal, and hence the `GU' keeps delaying the action execution. Ours, on the other hand, performs the task all 20 times as it models \textit{acceptable error space} using a multi-modal distribution. For the grasp success task `GU' baseline also has higher average views per decision compared to ours which shows that it unnecessarily postpones the decision. 

For objects like `mustard bottle' and `bleach cleanser' uni-modal pose distributions can often be obtained as they have unique texture on at least one side. However, acceptable errors in the pose estimate vary across different axes. As `GU' doesn't consider this, it often proceeds with the task execution and fails (a total of 9 failures for both objects). On the other hand, `Ours' considers the acceptable errors across different axes while making the decisions and as a result, has a significantly large number of successes and fewer failures compared to `GU' for both of these objects. 
% We further explain the results using qualitative examples for both tasks in Fig.~\ref{fig:base_qual} and \ref{fig:grasp_qual}.

% We don't see this trend with the availability of the IK solution task as often distribution incorrectly collapses to uni-modal distributions and as a result `Gaussian uncertainty' baseline attempts the task. However, there were no failures because the acceptable error for the base pose is very high. These results show that our method makes better decisions. We further provide two qualitative examples to justify this claim.

\subsection{Qualitative Results}
\begin{figure*}[]
    \centering
    \includegraphics[width=17.0cm]{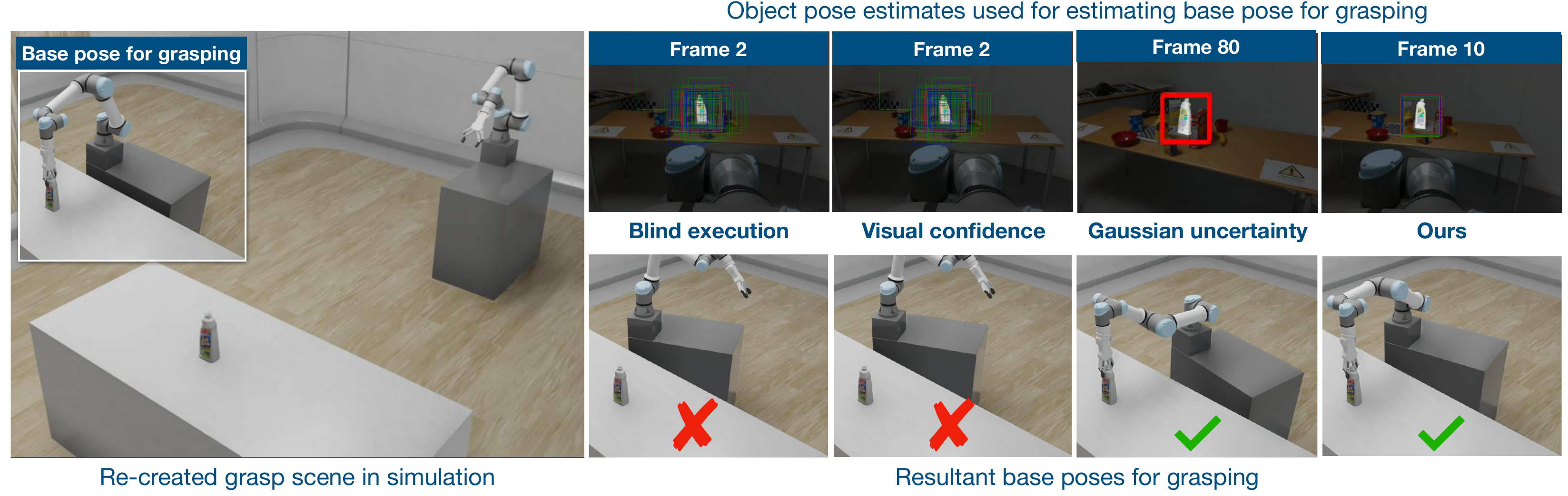}
    \caption{\textbf{Qualitative results for the availability of IK solution at the estimated base pose:} \textbf{Left}: Grasp scene and the selected base pose $\Tob$. \textbf{Right:} 
    Top row - Frames used by different baselines and our framework for performing the task and the estimated object poses $\poseestimate{}$. Bottom row - Estimated base poses for grasping $\baseposeest$.}
    \label{fig:base_qual}
\end{figure*}

In this section, we present two qualitative examples to further support our claims in Section~\ref{ss::ee:eob}. Fig.~\ref{fig:base_qual} presents an example for the availability of IK solution task. The `BE' chooses to navigate to the estimated base pose in frame 2 (as the base pose estimate in frame 1 would have resulted in a collision with the table). However, due to an error in the object pose estimate, there is also an error in the base pose estimate, and hence no IK solution is available for grasping the object at the estimated base pose. The `VC', similarly decides to navigate in frame 2 and encounters the same outcome. The `GU' defers the decision until frame 80, when the pose uncertainty has converged to a uni-modal distribution and the IK solution is available for grasping from the estimated base pose. In contrast, `Ours' decides to navigate as early as frame 10 and the IK solution is available for grasping the object from the estimated base pose.
\begin{figure*}[]
    \centering
    \includegraphics[width=17.0cm]{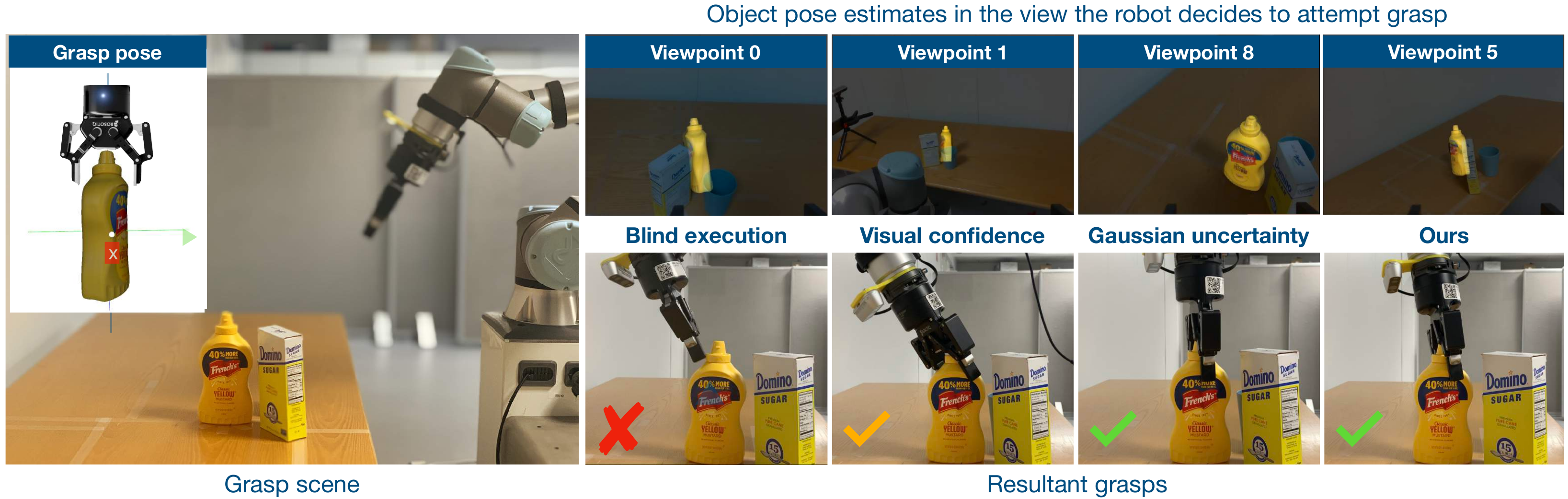}
    \caption{\textbf{Qualitative results for the grasp success using the estimated grasp pose:} \textbf{Left}: Grasp scene and selected grasp pose $\Tog$. \textbf{Right:} 
    Top row - Viewpoints used by different baselines and our framework for performing the task and the estimated object poses $\poseestimate{}$. Bottom row - Estimated grasp poses $\graspposeest$.}
    \label{fig:grasp_qual}
\end{figure*}

In Fig.~\ref{fig:grasp_qual}, we present an example for the grasp success task. The `BE' chooses to execute the grasp using the available pose estimate in the default viewpoint (viewpoint 0), despite severe occlusion of the object and a large error in the estimated object pose, which results in a failed grasp. The `VC' postpones the decision to viewpoint 1 due to a low confidence score caused by heavy occlusion in viewpoint 0. Although the object is less occluded in viewpoint 1 compared to viewpoint 0, no unique features are visible, leading to continued error in the object pose and the robot barely manages to grasp the object (unstable grasp). The `GU' defers the decision until the last viewpoint, when unique object features become visible and the pose uncertainty converges to a uni-modal distribution, allowing for a successful, stable grasp. In contrast, `Ours' opts to attempt the grasp using the pose estimate at viewpoint 5, where unique object features are just beginning to appear, resulting in a significant decrease in the object pose uncertainty and successfully completes the task.

\section{CONCLUSION AND FUTURE WORK}
\label{sec:con_and_fut}
In this work, we have introduced a framework to predict the likelihood of a robotic task's success, considering the uncertainty in estimated object poses and the acceptable error margins for task success. Our results show that by representing both the \textit{estimated error space} for the object pose estimate and the \textit{acceptable error space} for the task success using multi-modal distributions, we achieve 91\% success rate for the grasping task and 96\% success rate for the availability of IK solution task compared to 75\% and 90\% success rates respectively for the best-performing baselines. Further, we only have 6\% failures for the grasping task and no failures for the availability of IK solution as opposed to failure rates of 16\% and 6\%, respectively, for the best-performing baselines. As grasping is more sensitive to pose uncertainties, results are more significant for the grasping task. Further, by offloading the computationally expensive task of determining if the error is acceptable for the task through dynamic simulation, our framework achieves computational efficiency in online applications. Moreover, our proposed framework is generic and can be applied to a wide range of robotic tasks requiring object pose estimation. Hence, given the recent advancements in object pose uncertainty estimation and dynamic simulations, the proposed framework, in conjunction with these advancements, has the potential to enable robots to make reliable and informed decisions under pose uncertainty.

% Simulation always doesn't represent reality

\addtolength{\textheight}{-0cm}   % (12 default) This command serves to balance the column lengths

%%%%%%%%%%%%%%%%%%%%%%%%%%%%%%%%%%%%%%%%%%%%%%%%%%%%%%%%%%%%%%%%%%%%%%%%%%%%%%%%

%%%%%%%%%%%%%%%%%%%%%%%%%%%%%%%%%%%%%%%%%%%%%%%%%%%%%%%%%%%%%%%%%%%%%%%%%%%%%%%%



\section*{ACKNOWLEDGMENT}
This work was supported by the European Union’s Horizon 2020 project Fluently (grant agreement no 958417). The authors would also like to thank the SDU I4.0 lab for providing the robot for experimental evaluation and student Peter Khiem Duc Tinh Nguyen for helping with technical integration.

%%%%%%%%%%%%%%%%%%%%%%%%%%%%%%%%%%%%%%%%%%%%%%%%%%%%%%%%%%%%%%%%%%%%%%%%%%%%%%%%

\bibliographystyle{IEEEtran}
\typeout{}
\bibliography{references}

\end{document}